\documentclass[sigconf,nonacm]{acmart}
\renewcommand\footnotetextcopyrightpermission[1]{} 

\usepackage{fullpage}
\usepackage{hyperref}
\usepackage[compact]{titlesec}
\usepackage[linesnumbered,ruled,noend]{algorithm2e}
\SetKwInput{KwInput}{Input}
\SetKwInput{KwOutput}{Output}

\newcommand{\MYhref}[3][blue]{\href{#2}{\color{#1}{#3}}}%

   \def\E{\mathbb{E}}
  \def\P{\mathbb{P}}

\begin{document}

\fancyhead{}

\title{\textsc{ControlBurn}: Feature Selection by Sparse Forests }

\author{Brian Liu}
\affiliation{%
  \institution{Cornell University}
   \city{Ithaca}
   \state{New York}
  \country{USA}}
\email{bl462@cornell.edu}

\author{Miaolan Xie}
\affiliation{%
  \institution{Cornell University}
   \city{Ithaca}
   \state{New York}
  \country{USA}}
\email{mx229@cornell.edu}

\author{Madeleine Udell}
\affiliation{%
  \institution{Cornell University}
   \city{Ithaca}
   \state{New York}
  \country{USA}}
\email{udell@cornell.edu}

\maketitle

Tree ensembles distribute feature importance evenly amongst groups of correlated features. The average feature ranking of the correlated group is suppressed, which reduces interpretability and complicates feature selection. In this paper we present \textsc{ControlBurn}, a feature selection algorithm that uses a weighted LASSO-based feature selection method to prune unnecessary features from tree ensembles, just as low-intensity fire reduces overgrown vegetation. 
Like the linear LASSO, \textsc{ControlBurn} assigns all the feature importance of a correlated group of features to a single feature. 
Moreover, the algorithm is efficient and only requires a single training iteration to run,
unlike iterative wrapper-based feature selection methods.
We show that \textsc{ControlBurn} performs substantially better than feature selection methods with comparable computational costs on datasets with correlated features.

\begin{figure}[h]
     \includegraphics[width=\textwidth]{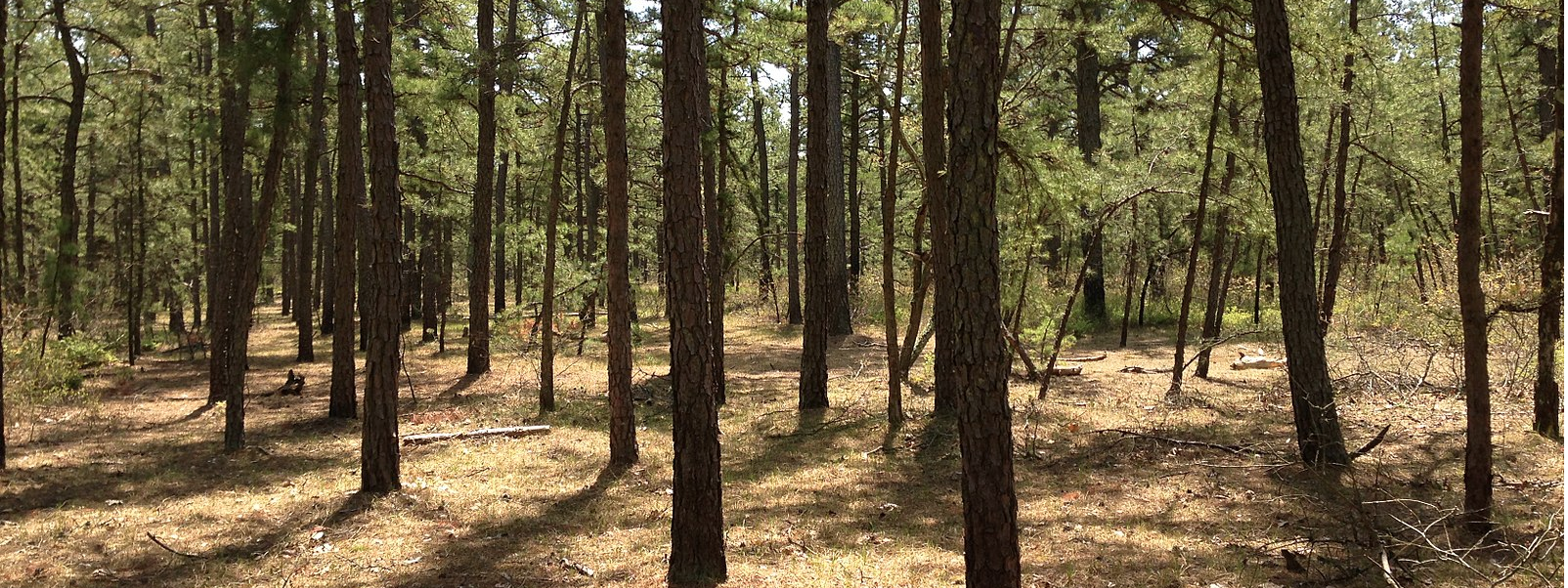}
     \caption{A sparse forest.}
   \end{figure}

\section{Introduction}

Tree ensembles are well-loved in machine learning for balancing excellent predictive performance with model interpretability. 
Importantly, tree ensemble models tend not to overfit 
even when the number of features is large.
And even complex boosting ensembles
admit feature importance scores that 
are cheap to compute. 
These properties make tree ensembles a popular choice 
for feature selection: simply train a model on all the features,
and select the most important features according to the 
feature importance score.

\begin{figure}[h]
\centerline{\includegraphics[width = 0.5\textwidth]{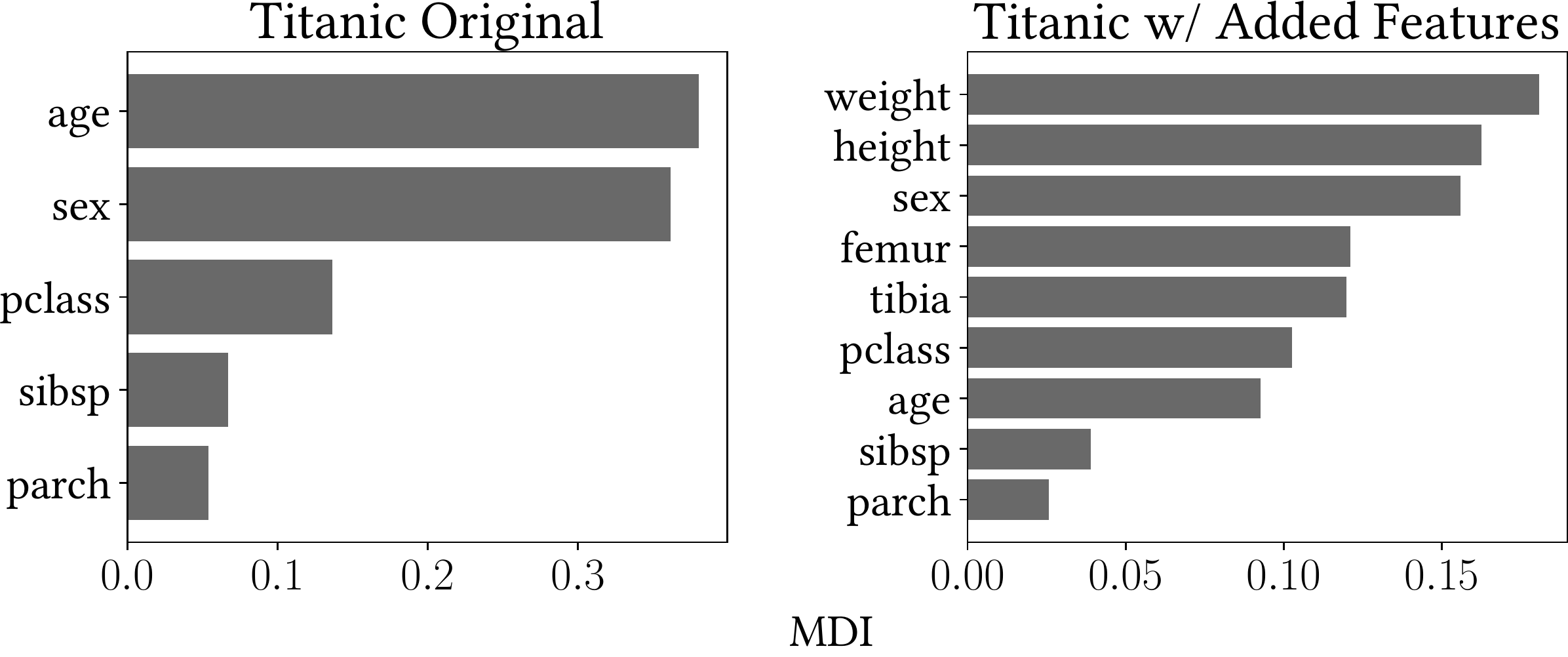}}
\caption{Effect of correlation bias on the top feature ranks in the Titanic dataset. The relative importance of "pclass" is inflated above "age".}
\label{titanic_bias.fig}
\end{figure}

However, when tree ensembles are trained on data with correlated features, 
the resulting feature rankings can be unreliable. 
This \emph{correlation bias} 
arises because tree ensembles split feature importance scores 
evenly amongst groups of correlated features. 
Figure \ref{titanic_bias.fig} shows an example. 
On the left plot, feature importance is calculated for a
random forest classifier fit on the Titanic dataset. 
The forest identifies age, sex, and passenger class 
as the most influential features. 
As an experiment, we modify the Titanic dataset 
by adding the synthetic features ``height'', ``weight'', ``femur'' length, and ``tibia'' length,
which we generate from a plausible distribution that 
depends on age and sex, 
following measurements from \cite{howell2000demography}.
 \thispagestyle{empty} 
As a result, ``height'', ``weight'', ``sex'', ``age'',  ``femur'', and ``tibia'' are highly correlated.
While these features together are important,
the feature importance of each variable in this cluster drops.
More generally, when large groups of correlated features are present in a dataset, 
important features may appear much less influential than they actually are.
Simple feature selection methods may select the whole group,
omit the entire group,
or pick the wrong features from the group. Here, ``weight'' and ``height'', synthetic noisy features,
are now elevated above more important features like ``age'' and ``sex''. In addition, ``pclass'' is elevated above ``age''. 


In this paper we develop \textsc{ControlBurn}, 
a feature selection algorithm that eliminates correlation bias by sparsifying the forest, 
just as fire is used in forest management to 
maintain a healthy (biological) forest by reducing density. 
We first grow a forest using techniques such as bagging and boosting. 
Then we reduce the number of trees by choosing sparse weights for each tree 
via a group-lasso penalized objective
that balances feature sparsity with predictive performance.

Importantly, our algorithm is efficient and scalable. We only need to fit a tree ensemble once and solve a group LASSO problem, which scales linearly with the number of data points, to select a sparse subset of features.

On our modified Titanic dataset, \textsc{ControlBurn} selects ``sex'', ``pclass'', and either ``age'' or a continuous proxy for age (``weight'', ``height'', ``femur'' or ``tibia'') as the three most important features. Since \textsc{ControlBurn} is robust to correlation bias, these selected features align with the three most important features in the original data.

\subsection{Applications}

\subsubsection{Interpretable machine learning}
Correlation bias reduces the interpretability of feature rankings. 
When features importance scores are split evenly amongst groups of correlated features, influential features are harder to detect:
feature ranks can become cluttered with many features that represent similar relationships. 
\textsc{ControlBurn} allows a data scientist to quickly identify a sparse subset of influential features for further analysis.

\subsubsection{Optimal experimental design}
In many experimental design problems, 
features are costly to observe. 
For example, in medical settings, certain tests are expensive and risky to the patient. 
In these situations it is crucial to avoid redundant tests. 
In this setting, \textsc{ControlBurn} can be used to identify
a subset of $k$ features (for any integer $k>0$) 
with (nearly the) best model performance,
eliminating redundancy.

\subsection{Main contributions}
The main contributions of this paper follow.
\begin{itemize}
    \item  We develop \textsc{ControlBurn}, a feature selection algorithm robust to correlation bias that can recover a sparse set of features from highly intercorrelated data.
    \item We present empirical evidence that \textsc{ControlBurn} outperforms traditional feature selection algorithms when applied to datasets with correlated features.

\end{itemize} 

\subsection{Paper organization}
We first examine related work concerning tree ensembles and feature selection techniques. Following this literature review, we explain in detail our implementation of \textsc{ControlBurn}. We then describe our experimental procedure to evaluate \textsc{ControlBurn} against traditional feature selection methods. Finally we present our findings and discuss interesting directions for future research. 

\section{Background}

\subsection{Tree ensembles}
The following section provides an overview of tree ensembles and details how feature importance is computed in tree models.

\subsubsection{Decision trees} form the base learners used in tree ensembles. We give a quick overview of classification trees below. A classification tree of depth $d$ separates a dataset into at most $2^d$ distinct partitions. Within each partition, all data points are predicted as the majority class. Classification trees are grown greedily, and splits are selected to minimize the Gini impurity, misclassification error, or deviance in the directly resulting partitions. The textbook \cite[Chapter 9]{friedman2001elements} provides a comprehensive reference on how to construct decision trees.

\subsubsection{Bagging} reduces the variance of tree ensembles through averaging the predictions of many trees. We sample the training data uniformly with replacement to create \emph{bags} (bootstrapped datasets with the same size as the original) and fit decision trees on each bag.
The final prediction is the average (or majority vote, for classification) of the predictions from all the trees. 
Reducing correlation between trees improves the effectiveness of bagging. 
To do so, we can restrict the number of features considered each time a tree is built or a split is selected \cite[Chapter 15]{friedman2001elements}.

\subsubsection{Boosting} fits an additive sequence of trees to reduce the bias of a tree ensemble. 
Each decision tree is fit on the residuals of its predecessor; 
weak learners are sequentially combined to improve performance. 
Popular boosting algorithms include  AdaBoost \cite{freund1999short}, gradient boosting  \cite{friedman2001greedy}, and XGBoost \cite{chen2016xgboost}. The textbook \cite[Chapter 10]{friedman2001elements} provides a comprehensive reference on boosting algorithms.

\subsubsection{Feature importance scores}
can be computed from decision trees by calculating the decrease in node impurity for each tree split. Gini index and entropy are two commonly used metrics to measure this impurity. Averaging this metric over all the splits for each feature yields that feature's importance score. This score is known as Mean Decrease Impurity (MDI); 
features with larger MDI scores are considered more important. MDI scores can be calculated for tree ensembles by 
averaging over all the trees in the ensemble \cite{2001.04295}.

\subsection{Feature selection}

Machine learning algorithms are sensitive to the underlying structure of the data. In datasets with linear relationships between the features and the response, linear regression, logistic regression, and support vector machines with linear kernels \cite{cortes1995support} are all suitable candidates for a model. For nonlinear datasets, tree ensembles such as random forests, gradient boosted trees, and explainable boosting machines (EBMs) \cite{nori2019interpretml} are commonly used. 

Feature selection is fairly straightforward in the linear setting. Methods such as LASSO \cite{tibshirani1996regression}, regularized logistic regression \cite{lee2006efficient}, and SVMs with no kernels \cite{lee2006efficient} can select sparse models. Our work aims to extend regularization techniques commonly used in linear models to the non-linear setting. Existing feature selection algorithms for nonlinear models can be  divided into the following categories: filter-based, wrapper-based, and embedded \cite{chandrashekar2014survey}.

\subsubsection{Filter-based}
\label{fs_review} feature selection algorithms remove features suspected to be irrelevant based on metrics such as correlation or chi-square similarity \cite{ambusaidi2016building, lee2011filter}. 
Features that are correlated with the response are preserved while uncorrelated features are removed. 
Groups of intercorrelated features correlated with the response are entirely preserved.

Filter-based algorithms are model-agnostic, their performance depends on the relevancy of the similarity metric used. For example, using Pearson's correlation to select features from highly non-linear data will yield poor results.

\subsubsection{Wrapper-based}
 feature selection algorithms iteratively select features through retraining the model across different feature subsets. \cite{mafarja2018whale,maldonado2009wrapper}.

For example, in recursive feature elimination (RFE) the model is initially trained on the full feature set. The features are ranked by feature importance scores and the least important feature is pruned. The model is retrained on the remaining features and the procedure is repeated until the desired number of features is reached. Other wrapper-based feature selection algorithms include sequential and stepwise feature selection. All of these algorithms are computationally expensive since the model must be retrained several times.

\subsubsection{Embedded}
feature selection algorithms select important features as the model is trained. \cite{langley1994selection,blum1997selection}. For example in LASSO regression, the regularization parameter $\lambda$ controls the number of coefficients forced to zero, i.e. the number of features excluded from the model. For tree ensembles, MDI feature importance scores are computed during the training process. These scores can be used to select important features, however this algorithm is highly susceptible to correlation bias since MDI feature importance is distributed evenly amongst correlated features. \textsc{ControlBurn} extends LASSO to the nonlinear setting and is an embedded feature selection algorithm robust to correlation bias.

\subsection{Sparsity vs. stability}
Like LASSO in the linear setting,
\textsc{ControlBurn} trades stability for sparsity. 
To build a sparse model on a datasets with correlated features, 
\textsc{ControlBurn} selects a subset of features to represent the group. 
The algorithm is useful for selecting an independent set of features, but the selected set is not stable.
Indeed, an arbitrarily small perturbation to the dataset
can lead the algorithm to select a different set of features when the features are highly correlated.

The tradeoff between sparsity and stability is fundamental to machine learning \cite{xu2011sparse}. Feature selection and explanation algorithms such as LASSO regression, best-subset selection \cite{bertsimas2016best}, and LIME \cite{ribeiro2016should} encourage sparsity, 
while algorithms such as ridge regression and SHAP \cite{lundberg2017unified} are more stable but not sparse. 
To achieve both sparsity and stability,
the dataset may need to be very large \cite{zhou2021}.

\section{\textsc{ControlBurn} algorithm}
\subsection{Overview}
The \textsc{ControlBurn} algorithm works as follows. 
Given feature matrix $X \in \mathbb{R}^{m \times p}$ and response $y \in \mathbb{R}^{m}$, 
we build a forest consisting of $n$ decision trees to predict $y$ from $X$. 
We will describe in detail in the next section 
various approaches to build this forest. 
Each tree $t_{i}$ for $i \in \{1,\ldots, n\}$ generates a vector of predictions $\alpha_{i} \in \mathbb{R}^{m}$. 
For each tree $t_{i}$, let $g_{i} \in \{0,1\}^{p}$ be a binary indicator vector, 
with $g_{ij} = 1$ if feature $j \in \{1, \ldots, p\}$ 
generates a split in tree $t_{i}$.

Fix a regularization parameter $\lambda > 0$. 
We introduce a non-negative variable $w \in \mathbb{R}^{n}$ 
to represent the weight of each tree in the forest;
a tree $t_i$ with $w_i = 0$ plays no role in prediction 
and can be pruned.
Let $A\in \mathbb{R}^{m\times n}$ be the matrix with vectors $\alpha_{i}$ as columns and 
let $\mathrm{G}\in \mathbb{R}^{p\times n}$ be the matrix with vectors $g_{i}$ as columns.
Hence $Aw \in \mathbb{R}^n$ gives the predictions of the forest 
on each instance given weights $w$,
while the elements of $Gw \in \mathbb{R}^p$ give 
the total weight of each feature in the forest.

We seek a feature-sparse forest (sparse $Gw$)
that yields good predictions.
To this end, define a loss function $L(A,w,y)$ for the prediction problem at hand, for example square loss for regression $ L(A,w,y) = \left\|y-\mathrm{~A} w\right\|_{2}^{2}$, or 
logistic or hinge loss for classification.

Inspired by the group LASSO \cite{tibshirani1996regression},
our objective is to minimize the loss together with a sparsifying regularizer, over all possible weights $w \geq 0$ for each tree:

\begin{equation}\label{eq-lasso}
    \begin{array}{ll}
    \mbox{minimize} & \frac{1}{m}L(A,w,y)+\lambda\left\|\mathrm{G} w\right\|_{1} \\
    \mbox{subject to} & w\geq 0,
    \end{array}
\end{equation}
Since $w_i\geq 0$ for all $i$, we equivalently express this problem as

\begin{equation}\label{eq-lasso}
    \begin{array}{ll}
    \mbox{minimize} & \frac{1}{m}L(A,w,y)+\lambda \sum_{i=1}^n u_i w_i \\
    \mbox{subject to} & w\geq 0,
    \end{array}
\end{equation}
where $u_i$ is the number of features used in tree $i$. 
This objective encourages sparse models, 
as the regularization penalty increases as more features are used in the ensemble.
The hyperparameter $\lambda$ controls the sparsity of the solution.

Let $w^\star \in \mathbb{R}^{m} $ denote the solution to Problem~\ref{eq-lasso}.  
A tree $i$ with a non-zero weight $w_i>0$ is used by the ensemble, while
a tree with weight $w_i=0$ is omitted.
We say feature $j$ was selected by the ensemble if 
$(Gw^\star)_j$ is non-zero.

The final step of our algorithm fits a new tree ensemble model on the selected features. 
This step improves solution quality, just as refitting a least squares regression model on features selected from LASSO regression reduces the bias of the final model \cite{chzhen2019lasso}.

Our algorithm operates in two major steps:
first we grow a forest, 
and then we sparsify the forest by solving Problem~\eqref{eq-lasso}.
We discuss each step in detail below.

 \begin{figure}[h]
\centerline{\includegraphics[scale =0.5]{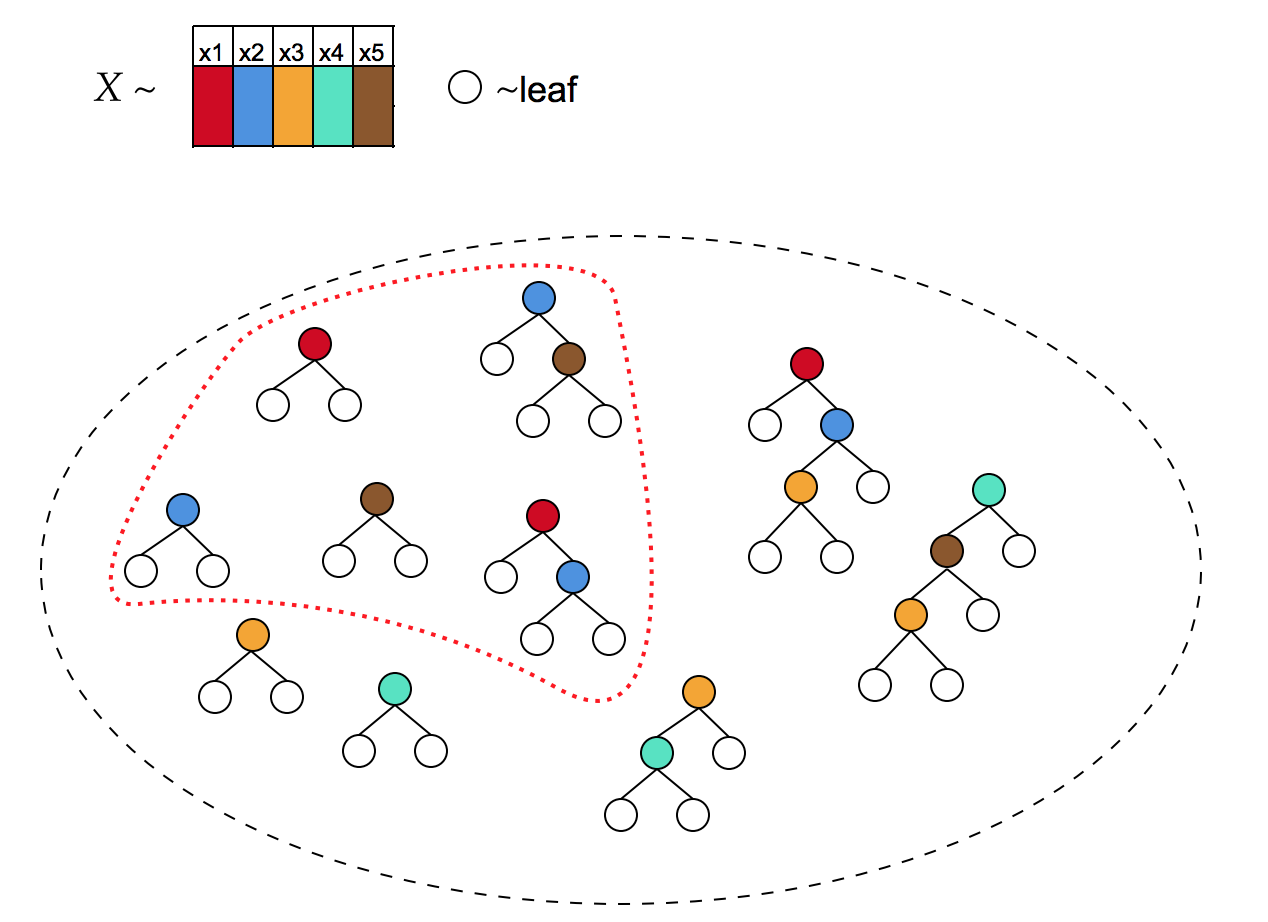}}
\caption{Selecting a feature sparse forest with \textsc{ControlBurn}.}
\end{figure}

\subsection{Growing a forest}

\textsc{ControlBurn} works best when executed on a diverse forest, with trees that use different sets of features. 
Traditional forest-growing methods such as bagging often fail to generate sufficient diversity. 
Using standard methods to grow the forest, 
we often find that no subset of the trees yields a 
feature-sparse model,
or that the resulting model has poor accuracy.

For example, if trees are grown to full depth, 
every tree in the forest generally uses the full feature set. 
Hence the only sparse solution $w^\star$ to Problem \ref{eq-lasso},
for any $\lambda$, is $w^\star = \mathbf{0}$: 
either all or none of the features will be selected.

We have developed two algorithms to grow diverse forests: incremental depth bagging and incremental depth bag-boosting.
Both are simple to tune and match the accuracy of other state-of-the-art tree ensembles.

\subsubsection{Incremental depth bagging}\label{s-inc-depth} (Algorithm \ref{incrementtree}) increases forest diversity by incrementally increasing the depth $d$ of trees grown. A tree of depth $d$ can use no more than $d^2 - 1$ features. As we increase $d$, the number of features used per tree increases,
allowing the model to represent more complex feature interactions. 

The algorithm has a single hyperparameter, $d^\text{max}$, the maximum depth of trees to grow. This hyperparameter is easy to choose: bagging prevents overfitting, so bigger is better, and $d^\text{max} = p$ is allowed.

We represent a tree ensemble as a set of trees $F = \{t_1 \ldots t_n $\}. With some abuse of notation, we let $F(x)$ be the output of the tree ensemble given input $x \in \mathbb{R}^p$.\\

\begin{algorithm}[H]
\DontPrintSemicolon
\KwInput{maximum depth $d^\textup{max}$}

Initialize $d\leftarrow 1$, $F \leftarrow \emptyset$ 

\While{$d \leq d^\textup{max}$}{
 
      Sample bag $X'$ from $X$ with replacement
     
      Fit tree $t$ of depth $d$ on $X',y$
     
      Add tree $t$ to forest: $F \leftarrow F \cup t$
    
     Compute train error of forest $F$
    
      If train error has converged \S\ref{s-inc-depth}, increment $d\leftarrow d+1$}

\KwOutput{Forest $F$}
\caption{Incremental Depth Bagging.}\label{incrementtree}
\end{algorithm}
 
 \begin{figure}[h]
\centerline{\includegraphics[scale=0.6]{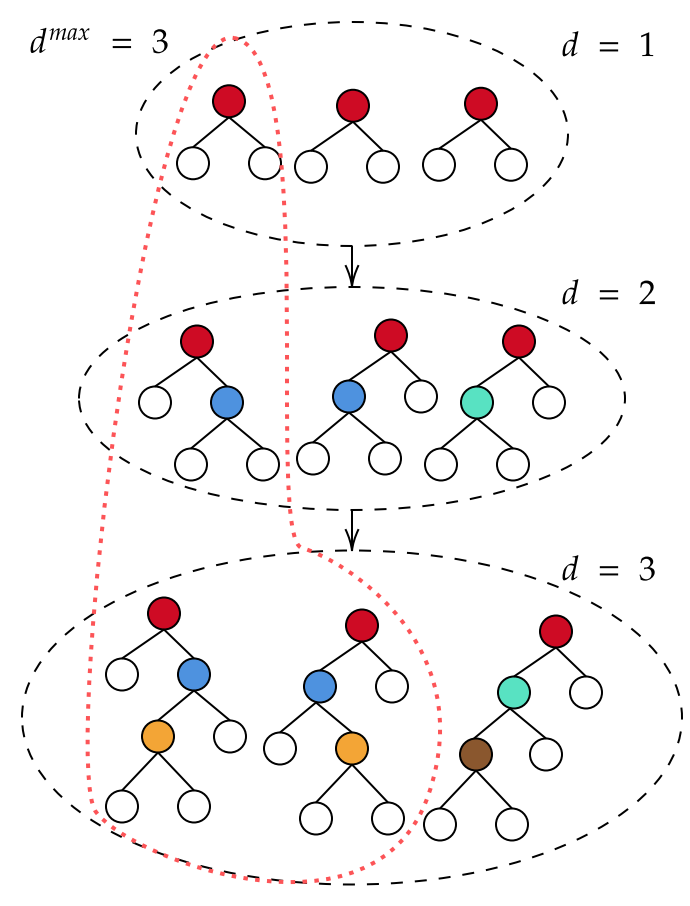}}
\caption{Incremental depth bagging with $d^\textup{max} = 3.$}
\label{baggingdiagram.fig}
\end{figure}

Incremental depth bagging increments the depth of trees grown when the performance of the model saturates at the lower depths. At each depth level, the training loss is recorded each time a new tree is added to the ensemble. When the training loss converges, the algorithm increments the depth $d$ and begins adding deeper trees. 
We say the loss \emph{converges} if the tail $N$ elements of the sequence fall within an $\epsilon$ tube. Experimentally, choosing $N = 5$ and $\epsilon = 10^{-3}$ yields good results.
This methodology differs from typical early stopping algorithms used in machine learning, since 
it stops based on training loss rather than test loss. 
We use training loss because it is fast to evaluate
and overfitting is not a concern,
since bagged tree ensembles tend not to overfit with respect to the number of trees used.

\subsubsection{Incremental depth bag-boosting}
(Algorithm \ref{bagboost}) 
uses out-of-bag (OOB) error to automatically determine the maximum depth of trees to grow, and requires no hyperparameters. The algorithm uses bagging ensembles as the base learner in gradient boosting to increase diversity without sacrificing accuracy.\\

\begin{algorithm}[H]
\DontPrintSemicolon
 Initialize $d\leftarrow 1$,  $F' \leftarrow \emptyset$, $F \leftarrow \emptyset$, $\delta > 0$ 

 Initialize $F(x)$ by predicting the mean/majority class of $y$, set $e \in \mathbb{R}^m$ as the vector of residuals

\While{$\delta > 0$ }{
  
     Sample bag $X'$ from $X$ with replacement
    
     Fit tree $t$ of depth $d$ on $X', e$
     
      $F' \leftarrow F' \cup t$
    
     Compute train error of forest $F \cup F'$
    
    \If{train error has converged \S\ref{s-inc-depth}} {
    
     $F \leftarrow F \cup F'$ 
    
     Set $e$ as negative gradient of loss

     Increment $d\leftarrow d+1$
     
     Set $\delta$, the improvement in OOB error from $F'$
     
     $F' \leftarrow \emptyset$
    
    }
}
\KwOutput{Forest $F$}
\caption{Incremental Depth Bag-Boosting.}\label{bagboost}
\end{algorithm}

  \begin{figure}[h]
\centerline{\includegraphics[scale =0.6]{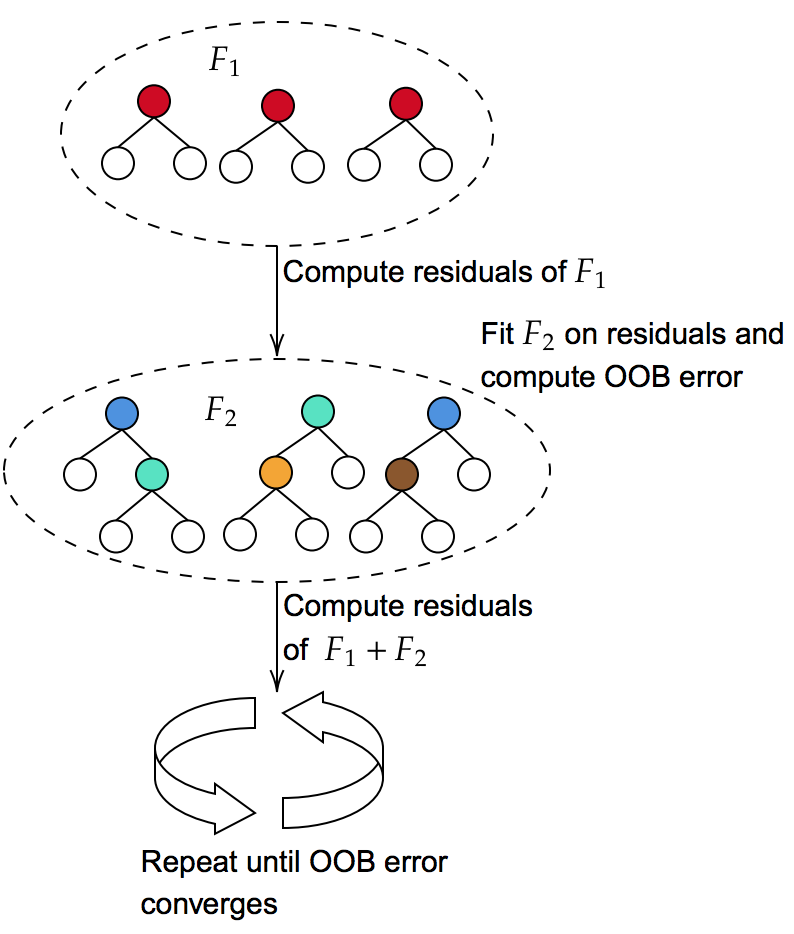}}
\caption{Iterations of incremental depth bag-boosting.}
\end{figure}

The number of trees grown per boosting iteration is determined by how fast the training error of the ensemble converges, as in \S\ref{s-inc-depth}. Each boosting iteration increments the depth of trees grown and updates the pseudo-residuals. This is done by setting $e_i = -\frac{\partial L(y_i, z)}{\partial z} \vert_{z = F(x_i)}$ for $i \in \{1\ldots m\}$.

For each row in $x \in X$, the OOB prediction is obtained by only using trees in $F^{'}$ fit on bags $\{X' \mid x \not \in X'\}$. Let $\delta$ be the improvement in OOB error between subsequent boosting iterations. The algorithm terminates when $\delta < 0$.

\begin{figure}[h]
\centerline{\includegraphics[scale=0.4]{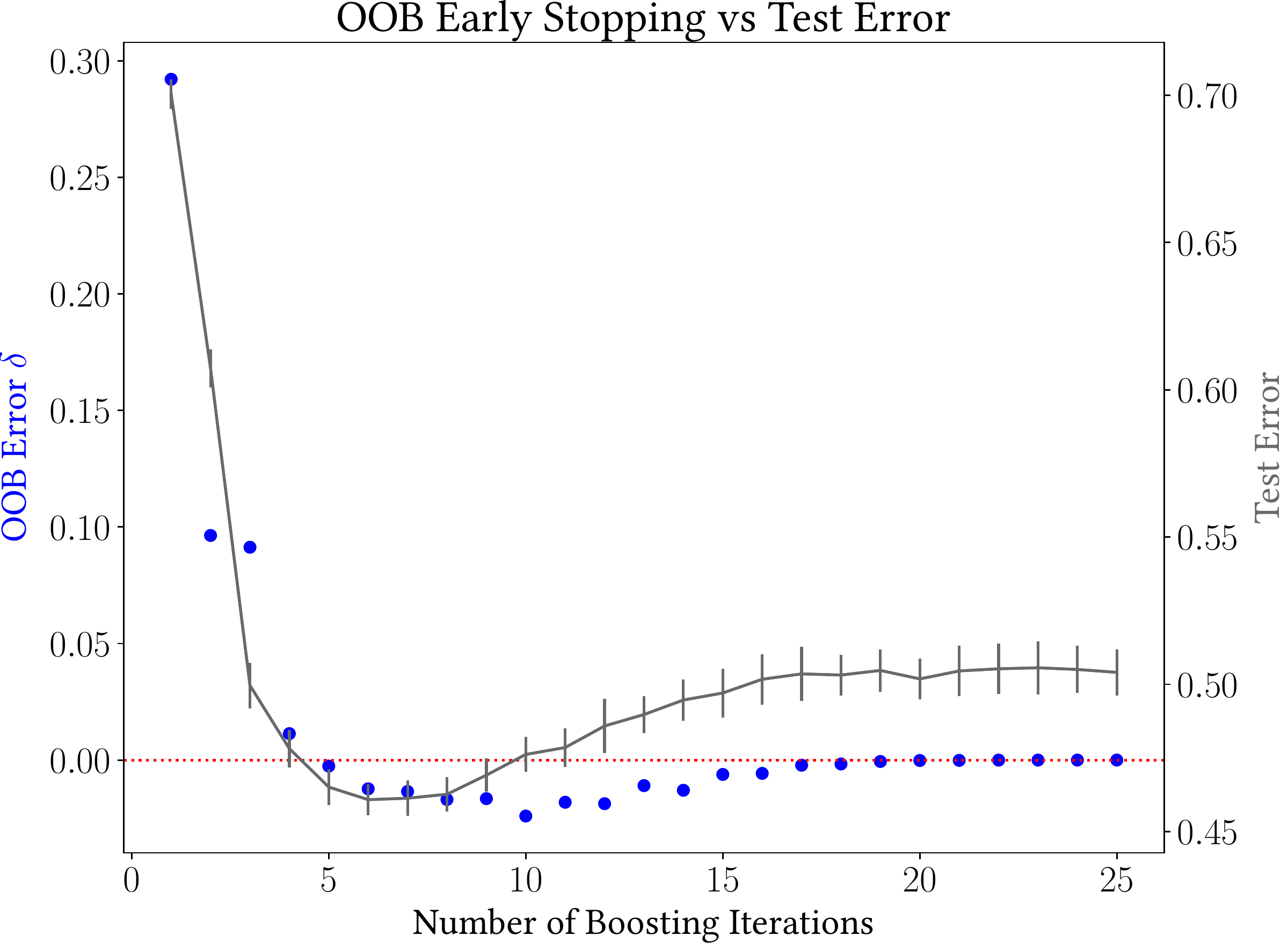}}
\caption{Incremental depth bag-boosting OOB early stopping vs. test error}
\label{OOB.fig}
\end{figure}

 In Figure \ref{OOB.fig}, incremental depth bag-boosting is conducted on the abalone dataset from UCI MLR \cite{Dua:2019}. The  scatterplot shows how the improvement in OOB error, $\delta$, changes as the number of boosting iterations increase. The procedure will terminate at the sixth boosting iteration, when $\delta < 0 $. The lineplot shows how out-of-sample (OOS) test error varies with the number of boosting iterations. OOS error is minimized at the seventh boosting iteration. This is remarkably close to the OOB error stop, which is determined on-the-fly.
 
\subsubsection{Theoretical guarantees}\label{s-theory}
The following lemma shows that the average performance improvement calculated using OOB samples is a good estimator of the true performance improvement for the data distribution. The lemma applies to any pair of ensembles $F$ and $\hat F$ 
where every element in ensemble $F$ appears in ensemble $\hat F$.

\begin{lemma}
Let $F_n$ be an ensemble with $n$ trees $\{T_1,\dots, T_n\}$, and let $F_{n+1}$ be an ensemble with one more tree $\{T_1,\dots, T_n, T_{n+1}\}$. Suppose the data (including the feature $x \in \mathbb{R}^p$ and label $y \in \mathbb{R}$) follows the distribution $(x,y) \sim \mathcal{X}$. 
Define the improvement $\textup{Imp}(x,y)=|y - F_n(x)|-|y - F_{n+1}(x)|$. 
Let $\bar I = \E_{(x,y)\sim \mathcal{X}}\textup{Imp}(x,y)$ be the expected improvement of the model $F_{n+1}$ over $F_n$, and $\mathcal{O}$ be the set of OOB samples of tree $T_{n+1}$. Suppose $\bar I$ is finite and each $(x,y)\in \mathcal{O}$ independently follows the distribution $\mathcal{X}$. Then 
$$\E\left(\frac{1}{|\mathcal{O}|}\sum_{(x,y)\in \mathcal{O}}\textup{Imp}(x,y)\right)= \bar I.$$ 

If the labels $y$ are all bounded, $\textup{Imp}(x,y)$ is a sub-exponential random variable with some parameters $(\nu,b)$, and for all $\epsilon\geq 0$,
\begin{align*}
    &\P\left(\left|\frac{1}{|\mathcal{O}|}\sum_{(x,y)\in \mathcal{O}}\textup{Imp}(x,y)-\bar I\right|\geq \epsilon \right)\leq 2e^{-\min\{\frac{\epsilon^2|\mathcal{O}|}{2\nu^2},\frac{\epsilon|\mathcal{O}|}{2b}\}}.
\end{align*}

\end{lemma}

We found empirically that incremental depth bag-boosting performs better than incremental depth bagging when used in \textsc{ControlBurn}. In addition, the OOB early stopping rule eliminates all hyperparameters and limits the size of the forest. This reduces computation time in the following optimization step. Forests grown by incremental depth bag-boosting are  more feature diverse than those grown via incremental depth bagging. Consider the case where a single highly influential binary feature dominates a dataset. While each iteration of incremental depth bagging will include this feature, the feature will only appear in the first boosting iteration in incremental depth bag-boosting. Boosting algorithms fit on the errors of previous iterations, so once the relationship between the feature and the outcome is learned, the feature may be ignored in subsequent iterations.

 \subsubsection{Other forest growing algorithms} include feature-sparse decision trees and explainable boosting machines (EBM).
 These algorithms do not perform as well as incremental depth bag-boosting, but may be more interpretable
 
 The procedure to build feature-sparse decision trees is a simple extension of the tradition tree building algorithm. For classification, decision trees typically choose splits to minimize the Gini impurity of the resulting nodes. Sparse decision trees add an additional cost $c$ to this impurity function when the tree splits on a feature that has not been used before. As a result, when groups of correlated features are present, the algorithm consistently splits on a single feature from each group. The cost parameter $c$ controls the sparsity of each tree. Ensembles of feature-sparse trees are not necessarily sparse, so \textsc{ControlBurn} must be used to obtain a final sparse model.
 
 Explainable boosting machines are generalized additive models composed of boosted trees trained on a single feature \cite{lou2013accurate}. Each tree in an EBM is feature-sparse with sparsity $1$. The additive nature of EBMs improve interpretability as the contribution of each feature in the model can be easily obtained. However, this comes at a cost of predictive performance since EBMs cannot capture higher order interactions. When \textsc{ControlBurn} is used on EBMs, the matrix $G \in \mathbb{R}^{p\times p}$ in the optimization step is the identity matrix.

\subsection{Optimization variants}

\subsubsection{Feature groupings}

In many applications, it may be necessary to restrict feature selection to a set of features rather than individual ones. For example, in medical diagnostics, a single blood test provides multiple features for a fixed cost. We can easily incorporate this feature grouping requirement into our framework by redefining $u_i$. Suppose features $1,\dots,p$ are partitioned into various groups. Instead of defining $u_i$ to be the number of features tree $i$ uses, we now define $$u_i=\sum_{ k\in T_k}c_k,$$ 
where $c_k$ is the cost of the group of features $ k$ and $T_k$ is the set of feature groups tree $i$ uses. Our original formulation corresponds to the case where all groups are singleton.

\subsubsection{Non-homogeneous feature costs}

In some problems the cost of obtaining one feature may be very different from the cost of obtaining another. We can easily accommodate this non-homogeneous feature costs requirement into our framework as well. Instead of defining $u_i$ to be the number of features used by tree $t_i$, we define $$u_i=\sum_{j\in T_i}c_j,$$ where $c_j$ is the cost of feature $j$ and $T_i$ is the set of features tree $i$ uses. Our original formulation corresponds to the case where all $c_j=1$.

\subsubsection{Sketching}
The optimization step of \textsc{ControlBurn} scales linearly with the number of rows in our data. We can sketch $X$ to improve runtime. Let $S\in \mathbb{R}^{s\times m}$ be the sampled (e.g., Gaussian) sketching matrix. Our optimization problem becomes:

\begin{equation}\label{eq-sketch}
\begin{array}{ll}
    \mbox{minimize} \quad &\frac{1}{s}\left\|Sy-\mathrm{~SA} w\right\|_{2}^{2}+\lambda \sum_{i=1}^n u_i w_i \\
    \mbox{subject to} & w\geq 0,
    \end{array}
\end{equation}


A relatively small $s$, around $O(\text{log}(m))$, can provide a good solution \cite{pilanci2015randomized}.

\section{Experimental setup}
We will evaluate how well \textsc{ControlBurn} selects features for binary classification problems.

\subsection{Data}
We report results on three types of datasets.

\subsubsection{Semi-synthetic}

To evaluate how well \textsc{ControlBurn} handles correlation bias, we synthetically add correlated features to existing data
by duplicating columns and adding Gaussian white noise to each replicated column. This procedure experimentally controls the degree of feature correlation in the data.

\subsubsection{Benchmark}
We consider a wide range of benchmark datasets selected from 
the Penn Machine Learning Benchmark Suite  \cite{olson2017pmlb}. This curated suite aggregates datasets from well-known sources such as the UCI Machine Learning Repository \cite{Dua:2019}. We select all datasets in this suite with a binary target, 43 datasets in total. 

\subsubsection{Case studies}
Finally, we select several real-world datasets known to contain correlated features, based on machine learning threads on Stack Overflow and Reddit. We select the Audit dataset, where risk scores are co-linear, and the Adult Income dataset, where age, gender, and marital status form a group of correlated features \cite{Dua:2019}. 
We also evaluate \textsc{ControlBurn} on the DNA microarray data used in West (2001) \cite{West:2001ft}, since gene expression levels are highly correlated.

\subsection{\textsc{ControlBurn} for experiments}

We first generate a forest of classification trees using incremental depth bag-boosting. The algorithm requires no hyperparameters and uses out-of-bag error to determine how many trees to grow. We use logistic loss as $L$ in optimization problem \ref{eq-lasso}. 
After solving the optimization problem, 
we eliminate all features with weight 0 and refit a random forest classifier on the remaining features. We evaluate the test ROC-AUC of this final model.\\

\begin{algorithm}[H]
\DontPrintSemicolon
\KwInput{desired sparsity $k$, 
maximum depth $d^\textup{max}$, initial regularization parameter $\lambda$, 
feature matrix $X$, labels $y$ }

Build forest $F$ with incremental depth bag-boosting (Algorithm~\ref{bagboost})

\Repeat{Desired sparsity of $F^*$ reached}{

 Use bisection \S\ref{s-bisection} to set $\lambda$ 
    
    Solve Problem~\ref{eq-lasso} using $\lambda$ to prune $F$
    
    Form sub-forest $F^*$ with trees of non-zero weights
}
\KwOutput{Forest $F^*$}
\caption{ControlBurn.}\label{controlburn}
\end{algorithm}

\subsubsection{Bisection}\label{s-bisection}
The regularization parameter $\lambda$ in Problem~\ref{eq-lasso} controls the number of features selected by \textsc{ControlBurn}, $k$. 
We use bisection to find the value of $\lambda$ that yields the desired number of selected features $k$. 
Our goal in feature selection is to produce an interpretable model, so we mandate that $k \leq k^\text{max} := 10$ in our experiments. 
Tree ensembles with more features can be difficult to interpret. 

In fact, we can use bisection with \textsc{ControlBurn} to 
generate feature sets of any sparsity between $0$ and $k^\text{max}$.
Start with a value of $\lambda$ large enough to guarantee $k = 0$.
Let $k^{'}$ denote the number of features we want \textsc{ControlBurn} to select in the following iteration and initialize $k^{'} = 1$. 
Bisect $\lambda = \frac{ \lambda}{2}$ and solve Problem~\ref{eq-lasso} to obtain $k$. 
If $k < k^{'}$, update  $\lambda = \frac{\lambda}{2}$; otherwise if $k > k^{'}$,  update $\lambda = \lambda + \frac{\lambda}{2}$. If $k = k^{'}$, update $k^{'} = k^{'} + 1$. 
Repeat until $k^{'} > k^\text{max}$.

\subsection{Baseline feature selection}

We will evaluate \textsc{ControlBurn} against the following baseline, an embedded feature selection method we call the \emph{random forest baseline}. 
Given feature matrix $X$ and response $y$, fit a random forest classifier and select the top $k$ features ranked by MDI feature importance scores. 
This feature selection algorithm is embedded since MDI feature importance scores are computed during training.

We fit a random forest classifier on the features selected by \textsc{ControlBurn} and the features selected by our random forest baseline and compare the difference in test performance. We use a 5-fold cross validation to estimate test error.

Our results in Section~\ref{s-experiments} omit comparisons between \textsc{ControlBurn} and filter-based and wrapper-based feature selection algorithms. We comment on comparisons here.

Experimentally, \textsc{ControlBurn} consistently outperforms filter-based algorithms across all datasets.

\begin{figure}[h]
 \centerline{\includegraphics[scale=.4]{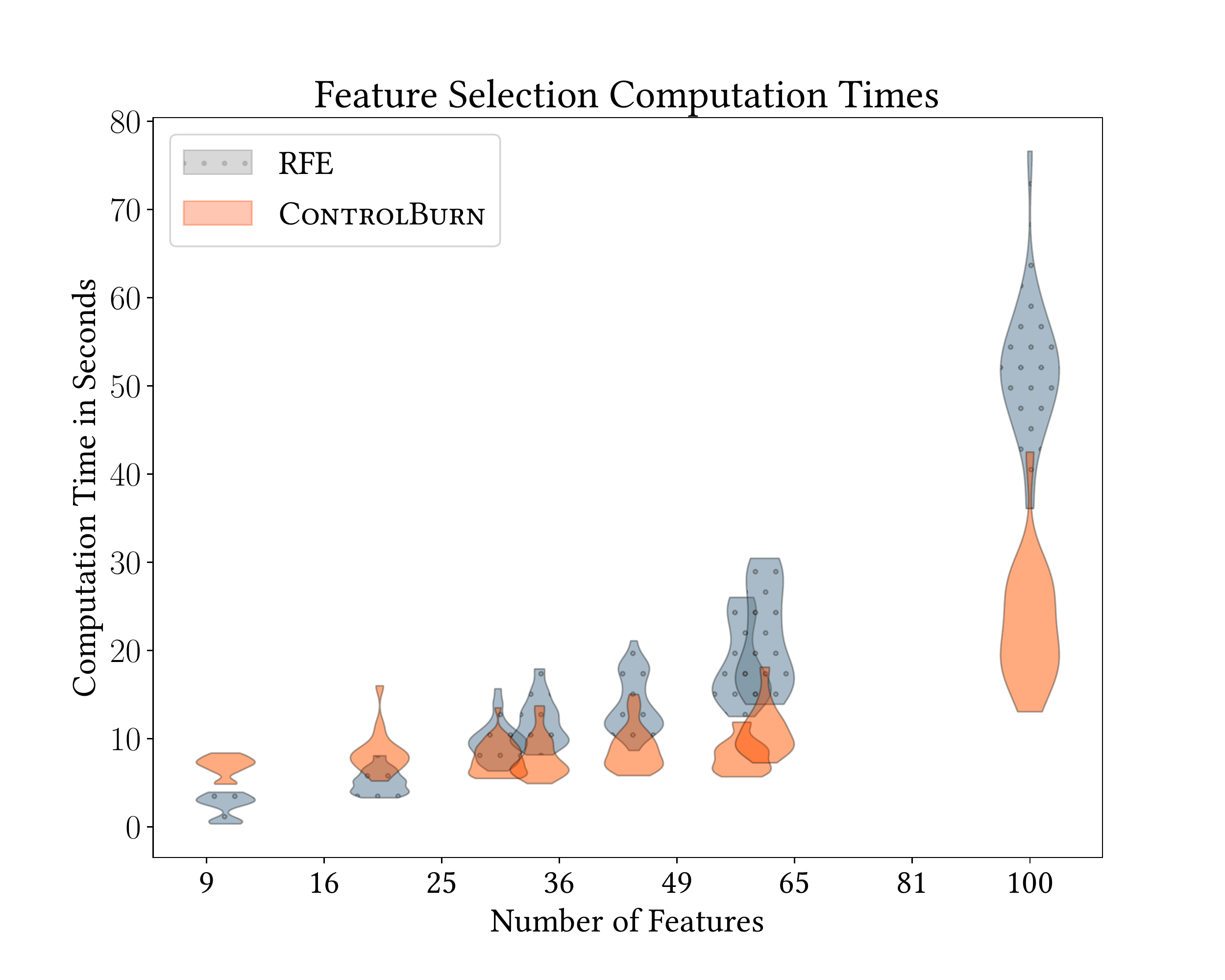}}

\caption{Computation time of RFE vs. \textsc{ControlBurn} as the number of features increase.}

\label{computetime.fig}
\end{figure}

Wrapper-based algorithms are computationally expensive compared to \textsc{ControlBurn} since the model must be retrained each time a new subset of features is evaluated. For example, selecting $k$ features using Recursive Feature Elimination (RFE) from a dataset with $p$ features requires $p-k$ training iterations. In Best Subset Selection, the number of iterations required scales exponentially with $p$.
In contrast, like all embedded feature selection algorithms, \textsc{ControlBurn} only requires a single training iteration to select $k$ features. To illustrate this advantage, we subsample our PMLB datasets to 1000 rows and compare computation time between \textsc{ControlBurn} and RFE. Figure \ref{computetime.fig} demonstrates that RFE runtime drastically increases with $p$ compared to \textsc{ControlBurn}.

\subsection{Implementation}
We use scikit-learn \cite{scikit-learn} to implement all random forest and decision tree classifiers in our experiment,
and we use CVXPY \cite{diamond2016cvxpy} and MOSEK \cite{mosek} to formulate and solve optimization problems. 

The code and data to reproduce our experiments can be found in \MYhref{https://github.com/udellgroup/controlburn}{this git repository}.

\section{Results}\label{s-experiments}

\textsc{ControlBurn} performs better than our random forest baseline on datasets that contain correlated features. 
In Figure~\ref{grid.fig}, we plot the performance of our selected models against the number of features selected by our two algorithms. On datasets with correlated features, such as the Cleveland Heart Disease dataset, \textsc{ControlBurn} performs better than the random forest baseline. On datasets where all of the features are independent, such as Breiman's synthetic TwoNorm dataset, there is no difference in performance.

In the following sections, we first discuss how \textsc{ControlBurn} performs when correlated features are added to semi-synthetic data. We then examine our results across our survey of PMLB datasets. Finally, we present several case studies that show the benefit of using \textsc{ControlBurn} on real world data.

\subsection{Semi-synthetic dataset}

\begin{figure}

    \centering    
    \includegraphics[scale=0.4]{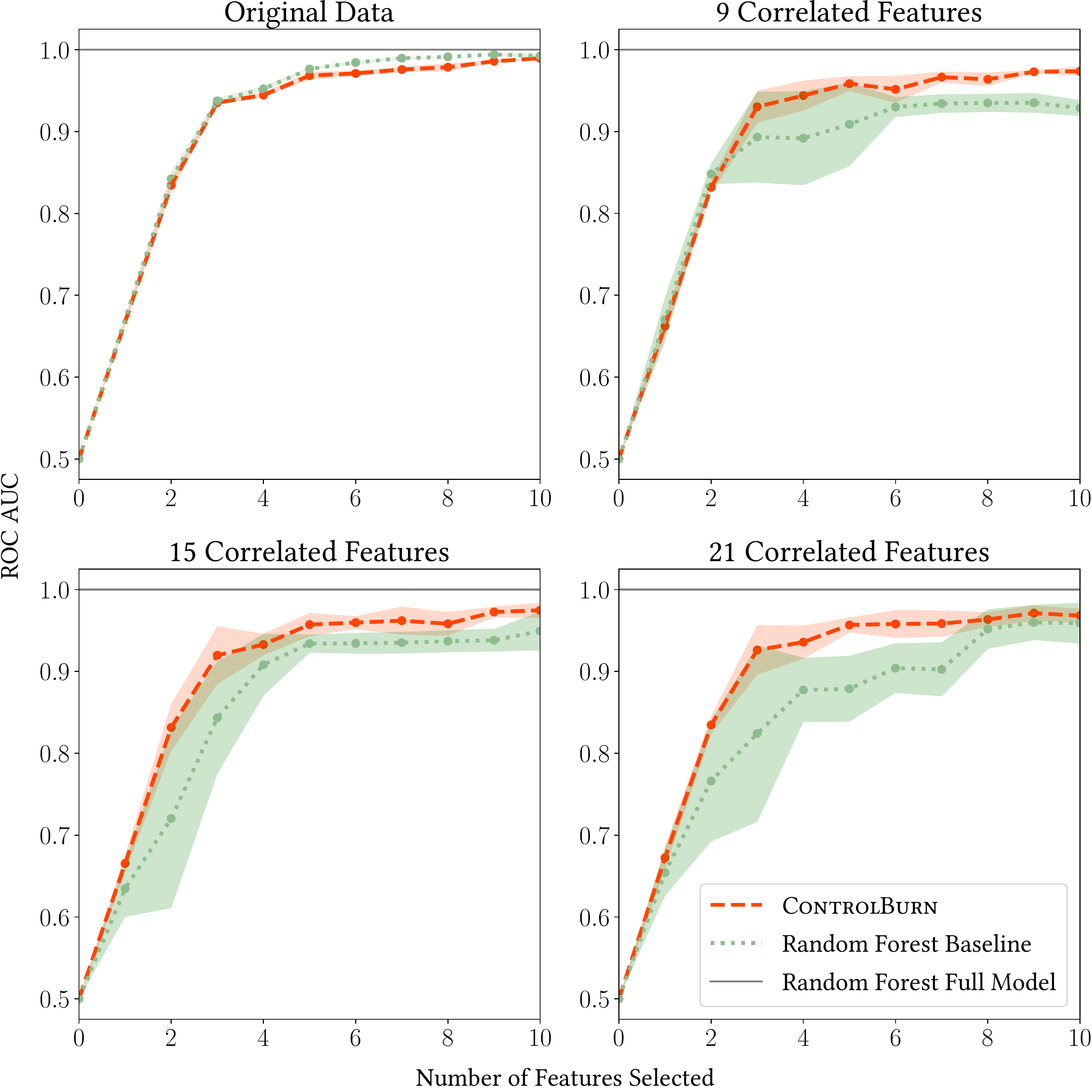}
    \caption{ \textsc{ControlBurn} performs well even with many highly correlated synthetic features. }
    
    \label{fig:Semi-Synthetic}
\end{figure}

To create our semi-synthetic dataset, we start with the Chess dataset from UCI MLR \cite{Dua:2019}. This dataset contains very few correlated features, since all features correspond to chessboard positions. We observe in figure \ref{fig:Semi-Synthetic} that \textsc{ControlBurn} performs nearly identically to our random forest baseline on the original data. We take the 3 most influential features, determined by random forest MDI importance, and create noisy copies by adding Gaussian noise with $\sigma = 0.1$. We repeat this process, duplicating the influential features 3, 5, and 7 times, which adds 9, 15, and 21 highly correlated features. Figure \ref{fig:Semi-Synthetic} shows that \textsc{ControlBurn} performs substantially better than the random forest baseline as the number of correlated features increases:
MDI feature importance is diluted amongst the groups of correlated features, depressing the ``importance'' rank of any feature in the correlated group. This semi-synthetic example illustrates that \textsc{ControlBurn} can effectively select features even when groups of strongly correlated features exist.

\subsection{PMLB suite}
\begin{figure}
    \centering    
\includegraphics[scale=0.4]{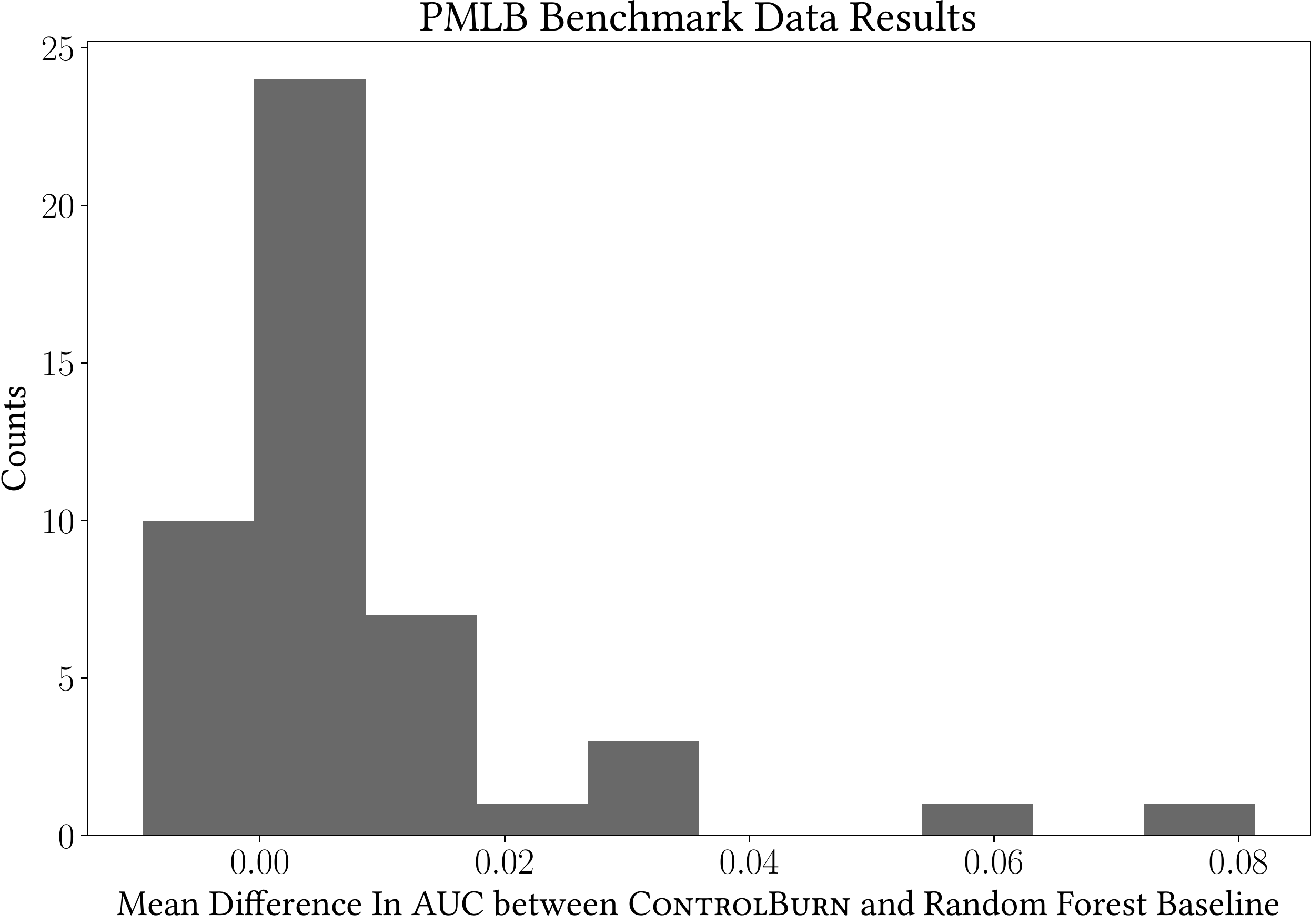}
    \caption{Differences in performance between \textsc{ControlBurn} and our random forest baseline.}
    \label{fig:pmlb}
\end{figure}

We evaluate \textsc{ControlBurn} on all $43$ binary classification datasets in the PMLB suite. On each dataset, we select $k$ features using \textsc{ControlBurn} and our random forest baseline and record the difference in ROC-AUC between the random forest classifiers fit on the selected features. 
We average these differences across sparsity levels $k \in \{1, \ldots, 10\}$ and present the distribution as Figure \ref{fig:pmlb}.

In this histogram, we see that \textsc{ControlBurn}
performs similarly to the random forest baseline for most of these datasets: most of the distribution is supported on $[-0.01,0.01]$, since most of the datasets in PMLB do not contain correlated features. 
Still, \textsc{ControlBurn} performs better than our random forest baseline on average: 
the mean of the distribution is positive, and the 
distribution is right-skewed. Most of the datasets along the right tail, such as Musk, German, and Sonar, contain correlated features.

\subsection{Case studies}

\begin{figure*}[h]
\centerline{\includegraphics[scale = 0.35]{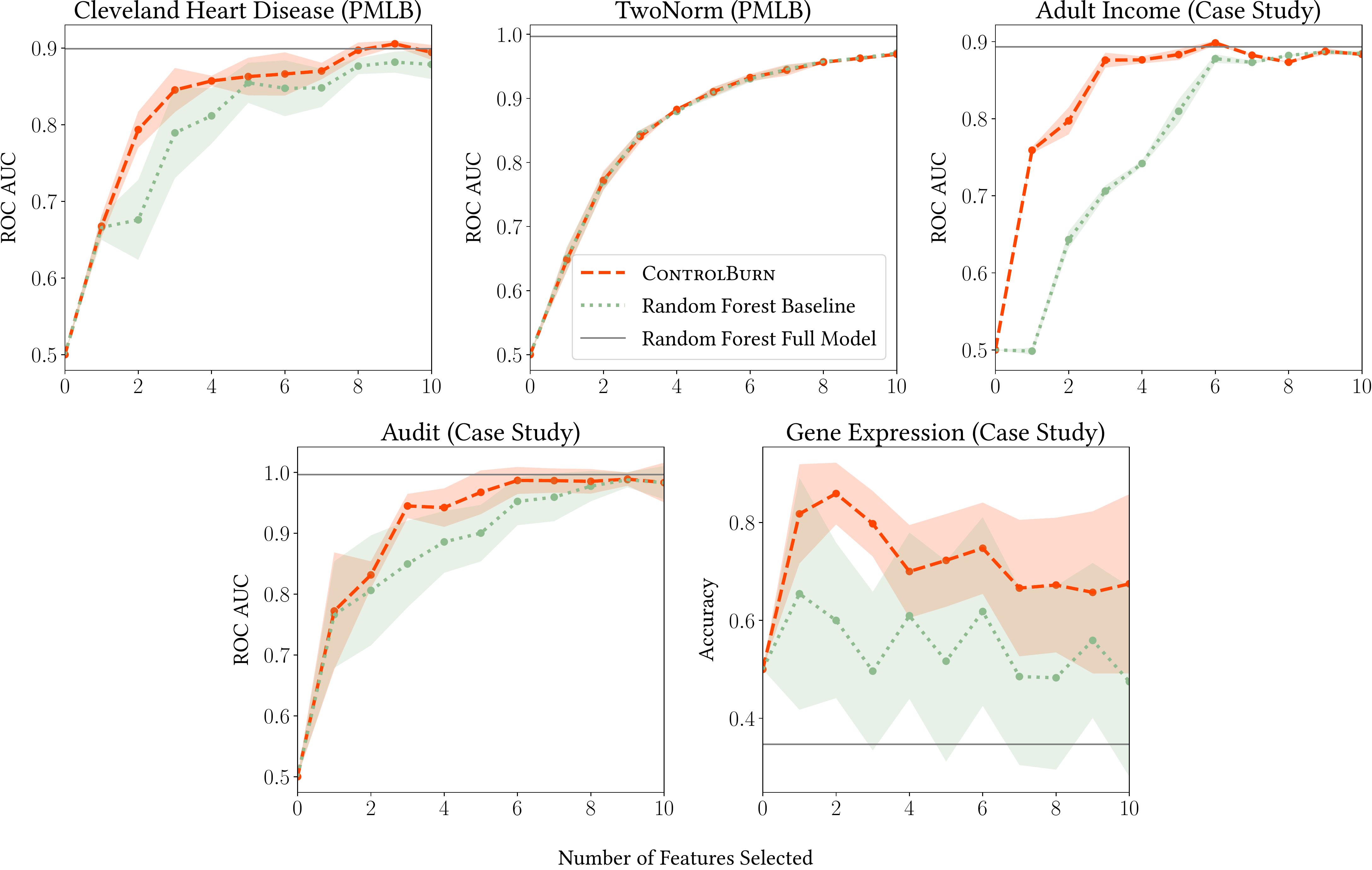}}
\caption{ The performance of \textsc{ControlBurn} and our random forest baseline plotted against the number of features selected.
\textsc{ControlBurn} always matches or exceeds the performance of the baseline (up to statistical error).}
\label{grid.fig}
\end{figure*}

\subsubsection{Adult income dataset}

We use \textsc{ControlBurn} to build a classification model to predict whether an adult's annual income is above \$50k. The Adult Income plot in figure \ref{grid.fig} shows how well \textsc{ControlBurn} performs compared to our random forest baseline. 
We see that \textsc{ControlBurn} performs substantially better than our baseline when $k = 3$ features are selected. \textsc{ControlBurn} selects ``EducationNum'', ``CapitalGain'', and ``MaritalStatus\_Married-civ-spouse'' while our random forest baseline selects ``fnlwgt'', ``Age'' and ``CapitalGain''.

\begin{figure}[h]
\centerline{\includegraphics[width = .65\textwidth]{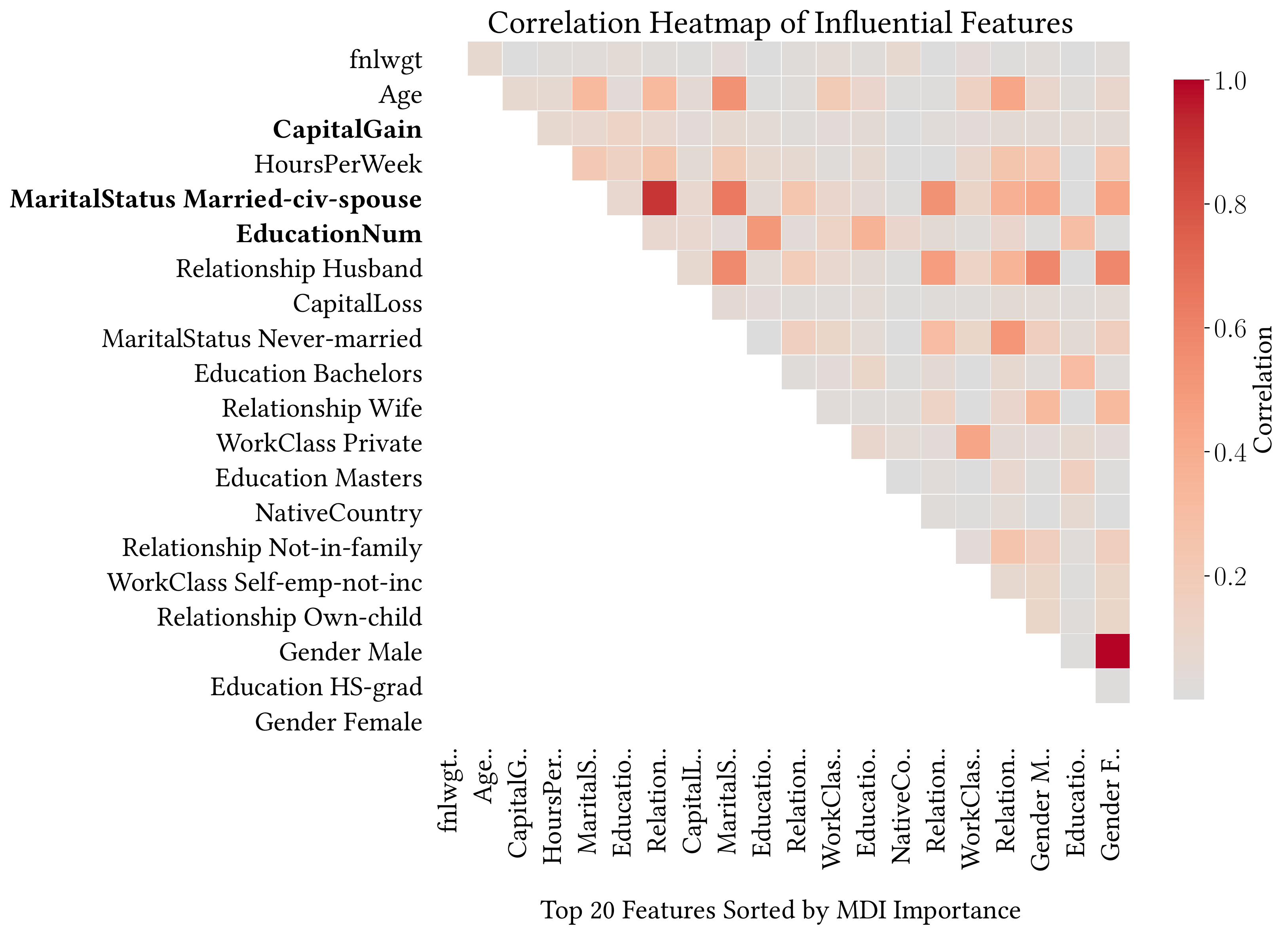}}
\caption{Adult Income Dataset: Top 20 features sorted by MDI importance. The top 3 features selected by \textsc{ControlBurn} are in bold.}
\label{figure_cor}
\end{figure}

Figure \ref{figure_cor} shows a correlation heat map of the top 20 features in the Adult Income dataset ranked by MDI importance. The top 3 features selected by \textsc{ControlBurn}  are in bold. From this plot, it is apparent that the MDI feature importance of  ``MaritalStatus\_Married-civ-spouse'' is diluted because the feature is correlated with many other influential features. \textsc{ControlBurn} is able to select this feature while our random forest baseline ignores it. 

It is also interesting to note that the feature ``fnlwgt'' is uninformative: it represents the population size of a row and is unrelated to income. This feature is given high importance by our random forest baseline but ignored by \textsc{ControlBurn}. We discuss this phenomenon further in Section \ref{split bias}.

\subsubsection{Audit dataset}
The Audit dataset demonstrates that highly correlated and even duplicated features sometimes appear in real-world data. Many of these risk scores are highly correlated and several scores are linear transformations of existing scores. We use \textsc{ControlBurn} to build a classification model to predict whether a firm is fradulent based on risk scores. Figure \ref{grid.fig} shows that \textsc{ControlBurn} outperforms our baseline for any sparsity level $k$.

\subsubsection{Microarray gene expression}

In this experiment, we use \textsc{ControlBurn} to determine which gene expressions are the most influential towards predicting the hormone receptor status of breast cancer tumors. Our data contains 49 tumor samples and over 7000 distinct gene expressions. We formulate this task as a binary classification problem to predict receptor status using gene expressions as features.  
The Gene Expression plot in Figure \ref{grid.fig} shows the difference in accuracy between a model selected using \textsc{ControlBurn} and our random forest baseline. \textsc{ControlBurn} outperforms our baseline for every sparsity $k$. 
These preliminary findings suggest that \textsc{ControlBurn} is effective at finding a small number of genes with important effects on tumor growth.
We discuss implications and extensions appropriate for 
gene expression data further in Section \ref{pbiggerm}. 

\section{Discussion and future work}

\begin{figure}[h]
\centerline{\includegraphics[width = 0.6 \textwidth]{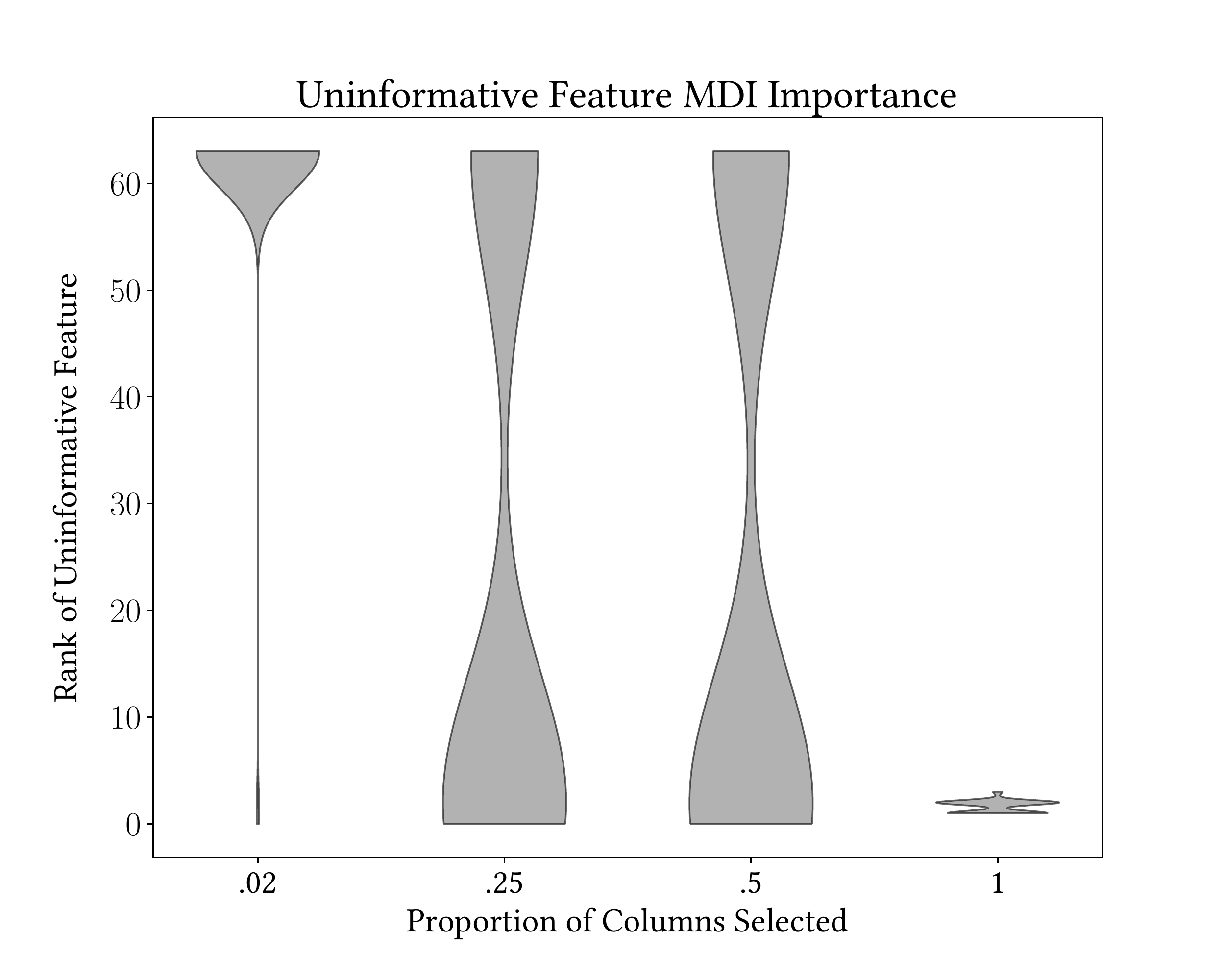}}
\caption{Rank of uninformative continuous feature vs. the proportion of features selected by \textsc{ControlBurn}.}
\label{splitbias.fig}
\end{figure}

\subsection{Bias from uninformative continuous features}
\label{split bias}

Brieman et. al (1984) \cite{brieman1984classification} and Strobl et. al (2007) \cite{strobl2007bias} note that MDI feature importance is biased towards features with more potential splits. The feature importance of continuous features, which can be split on multiple values, is inflated compared to binary features. 

We attempt to quantify whether \textsc{ControlBurn} suffers bias from uninformative continuous features. 
We replicate the experiment conducted by Zhou and Hooker (2019) \cite{zhou2019unbiased} by adding a random continuous feature to a classification dataset. We use \textsc{ControlBurn} to select a relevant subset of features and compute the feature rank, via MDI importance, of the uninformative feature in the model. Figure \ref{splitbias.fig} shows that when the full feature set is selected, the uninformative feature is often present in the top feature rankings due to bias.
As we increase $\lambda$ and select fewer features, 
our uninformative feature is demoted. 
This experiment demonstrates that \textsc{ControlBurn} mitigates bias from uninformative continuous features. It would be interesting to compare \textsc{ControlBurn} against other feature importance algorithms robust to this bias, such as Unbiased Feature Importance \cite{zhou2019unbiased}.

\subsection{Limits of LASSO: big $p$, small $m$}
\label{pbiggerm}
Zou and Hastie (2005) \cite{zou2005regularization} note a limitation of LASSO:
the number of features selected by LASSO is bounded by the number of rows $m$.  When \textsc{ControlBurn} is used for a problem with more features than rows ($p \gg m$), the number of nonzero entries in $w^\star$ is bounded by $m$. At most $m$ trees can be included in the selected model. If all trees are grown to full depth, the selected model can use at most $m^2$ features. 
In our gene expression example, $m = 49$ while $p > 7000$. 
The upper-bound on the number of features that can selected via \textsc{ControlBurn} is much higher than the corresponding upper-bound for LASSO regression, however \textsc{ControlBurn} may still be unable to select the entire feature set if $p > m^2$. \textsc{ControlBurn} also selects a single feature from a group of correlated features. This makes \textsc{ControlBurn} robust to correlation bias but ineffective for gene selection, where it is desirable to include entire groups of correlated gene expressions. 

We propose extending \textsc{ControlBurn} by adding an $\ell_2$ regularization term to our optimization problem, just as the elastic net extends LASSO regression in the linear setting. 
Group elastic nets for gene selection have been explored in Munch (2018) \cite{munch2018adaptive}; using this approach to select tree ensembles presents an interesting direction for future research.

\section{Conclusion}
In this work, we develop \textsc{ControlBurn}, a weighted LASSO-based feature selection algorithm for tree ensembles. Our algorithm is robust to correlation bias; when presented with a group of correlated features, \textsc{ControlBurn} will assign all importance to a single feature to represent the group. This improves the interpretability of the selected model, as a sparse subset of features can be recovered for further analysis. \textsc{ControlBurn} is flexible and can incorporate various feature groupings and costs. In addition, the algorithm is computationally efficient and scalable.

Our results indicate that \textsc{ControlBurn} performs better than traditional embedded feature selection algorithms across a wide variety of datasets. We also found our algorithm to be robust to bias from uninformative continuous features. \textsc{ControlBurn} is an effective and useful algorithm for selecting important features from correlated data.
\bibliographystyle{acm} 
\bibliography{reference}

\newpage
\appendix
\section{Benchmark Datasets}

\begin{table}[h]

\label{pmlbtable}
\begin{tabular}{|l|l|l|}
\hline

\textbf{dataset}             & \textbf{Rows} & \textbf{Columns} \\ \hline
analcatdata\_cyyoung8092     & 97                    & 10                   \\ \hline
australian                   & 690                   & 14                   \\ \hline
biomed                       & 209                   & 8                    \\ \hline
breast\_cancer\_wisconsin    & 569                   & 30                   \\ \hline
churn                        & 5000                  & 20                   \\ \hline
cleve                        & 303                   & 13                   \\ \hline
colic                        & 368                   & 22                   \\ \hline
diabetes                     & 768                   & 8                    \\ \hline
glass2                       & 163                   & 9                    \\ \hline
Hill\_Valley\_without\_noise & 1212                  & 100                  \\ \hline
horse\_colic                 & 368                   & 22                   \\ \hline
ionosphere                   & 351                   & 34                   \\ \hline
ring                         & 7400                  & 20                   \\ \hline
spambase                     & 4601                  & 57                   \\ \hline
tokyo1                       & 959                   & 44                   \\ \hline
\end{tabular}
\vspace{3mm}
\caption{Sample of PMLB benchmark datasets with dimensions.}
\end{table}

\end{document}